\documentclass[runningheads]{llncs}

\usepackage{eccv}

\usepackage{eccvabbrv}

\usepackage{graphicx}
\usepackage{booktabs}

\usepackage{xcolor} 
\usepackage{makecell} 
\usepackage{colortbl}
\usepackage{multirow}
\usepackage{diagbox}
\usepackage{algorithm}
\usepackage{algorithmic}
\usepackage{color}

\usepackage[accsupp]{axessibility}  

\usepackage{hyperref}

\usepackage{orcidlink}

\makeatletter
\def\thanks#1{\protected@xdef\@thanks{\@thanks
        \protect\footnotetext{#1}}}
\makeatother

\hypersetup{
    colorlinks=true
}

\begin{document}

\title{Bidirectional Uncertainty-Based Active Learning for Open-Set Annotation} 

\titlerunning{BUAL for Open-Set Annotation}

\author{Chen-Chen Zong\inst{1}\orcidlink{0000-0003-3588-1461} \and
Ye-Wen Wang\inst{1}\orcidlink{0009-0004-4147-5341} \and
Kun-Peng Ning\inst{2}\orcidlink{0009-0006-8053-4310} \and
Hai-Bo Ye\inst{1}\orcidlink{0000-0001-9034-7013} \and
Sheng-Jun Huang\inst{1}†\thanks{† Corresponding Author: huangsj@nuaa.edu.cn}\orcidlink{0000-0002-7673-5367}}

\authorrunning{C.~Zong et al.}

\institute{Nanjing University of Aeronautics and Astronautics
\and Peking University}


\maketitle

\begin{abstract}
   Active learning (AL) in open set scenarios presents a novel challenge of identifying the most valuable examples in an unlabeled data pool that comprises data from both known and unknown classes. Traditional methods prioritize selecting informative examples with low confidence, with the risk of mistakenly selecting unknown-class examples with similarly low confidence. Recent methods favor the most probable known-class examples, with the risk of picking simple already mastered examples. In this paper, we attempt to query examples that are both likely from known classes and highly informative, and propose a \textit{Bidirectional Uncertainty-based Active Learning} (BUAL) framework. Specifically, we achieve this by first pushing the unknown class examples toward regions with high-confidence predictions, \textit{i.e.}, the proposed \textit{Random Label Negative Learning} method. Then, we propose a \textit{Bidirectional Uncertainty sampling} strategy by jointly estimating uncertainty posed by both positive and negative learning to perform consistent and stable sampling. BUAL successfully extends existing uncertainty-based AL methods to complex open-set scenarios. Extensive experiments on multiple datasets with varying openness demonstrate that BUAL achieves state-of-the-art performance. The code is available at this \href{https://github.com/chenchenzong/BUAL}{link}.
\end{abstract}

\section{Introduction}
\label{sec:intro}

Labeling data can be costly and time-consuming, often requiring high levels of expertise from annotators~\cite{settles2009active}. This expense poses a significant challenge when dealing with insufficient labeled data in deep learning tasks. Recently, active learning (AL) has emerged as a prominent approach to tackle this issue and has gained widespread attention~\cite{mahmood2021low,huang2021asynchronous,ren2021survey}. It iteratively selects the most informative examples from the unlabeled data pool and queries their labels from an oracle, enabling the learning of an effective model with reduced labeling costs.

Existing AL methods~\cite{roy2001toward,fu2013survey,huang2010active,you2014diverse,sinha2019variational,yoo2019learning,ning2021improving} typically operate under the closed-set assumption, assuming that the label categories in the unlabeled data pool match those of the target task. However, this assumption often does not hold in practical scenarios. For example, consider a task that involves classifying images into two target categories, "Dog" and "Cat". Collecting training examples through keyword-based image search inevitably introduces irrelevant images from other categories (\ie, unknown classes), alongside the two target categories (\ie, known classes).

In such open-set scenarios, many previous AL methods, which prefer querying examples with less confident predictions, may lead to failure since examples from unknown classes often receive uncertain predictions. To mitigate the impact of unknown class examples, some AL methods designed specifically for open-set scenarios attempt to query examples that are more likely to belong to known classes based on sample similarity~\cite{du2021contrastive} or model-predicted max activation value (MAV)~\cite{ning2022active}. However, since examples similar to the labeled ones may be already mastered by the model, they do not significantly benefit the target model. These methods only perform well when the proportion of unknown class examples is high. When the proportion is low, they tend to perform poorly than traditional AL methods, even inferior to random sampling (please refer to Figure \ref{fig.acc}). However, determining the proportion of unknown class examples in practical scenarios is often challenging. This further limits the usability of these methods.

In this paper, we start with a specific question: \textbf{can we effectively distinguish the "informative" examples of known classes from examples of unknown classes?} Intuitively, if we can push the unknown class examples toward regions with high-confidence predictions, existing uncertainty-based AL methods can be applied directly in open-set scenarios. To achieve this, we propose to fine-tune the model by performing negative learning (NL)~\cite{kim2019nlnl,luo2021exploiting,lee2021deep,zong2022noise} on unlabeled examples. NL is an indirect learning manner that explores the utility of complementary labels, \ie, the label categories that an instance does not belong to. For a $K$-classification problem, the NL loss is defined as:
\begin{equation}\label{eq1}
	\ell_{NL} \left ( f,\bar{y}  \right ) = -\textstyle\sum_{k = 1}^{K} \bar{y} _{k} \log_{}{(1-p_k)} .
\end{equation}
where $f$ is the model we want to optimize, $\bar{y}$ and $\boldsymbol{\bar{y}}=\left [ \bar{y}_1,\dots,\bar{y}_k,\dots,\bar{y}_K \right ] $ represent a complementary label and its corresponding one-hot form, respectively, and $p_k$ denotes the probability of the $k$-th category.

Specifically, the fine-tuning process comprises two parts. On one hand, for already labeled examples, we train them directly using Equation \ref{eq1}. On the other hand, for unlabeled examples, we first randomly assign labels to them in each training round and then train the model using Equation \ref{eq1}. Notably, the unlabeled known class example has a relatively higher chance of receiving the correct label, whereas the unknown class example will never be assigned the correct label. Once unlabeled known class data receive the correct labels, they suffer a larger penalty and are reduced confidence predictions by the model since they deviate from the distribution information obtained from labeled data. In contrast, unlabeled unknown class data are not constrained to move towards the high-confidence region for counteracting the update gradient produced by the labeled data.

\begin{figure}[tb]
	\centering
	\begin{subfigure}{0.475\linewidth}
		\includegraphics[width=1.\linewidth]{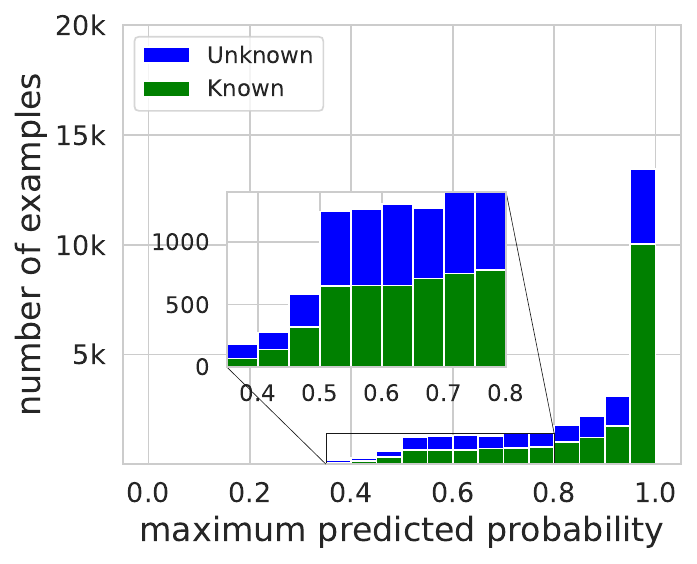}
		\caption{w/o NL}
		\label{fig.2a}
	\end{subfigure}
	\hfill
	\begin{subfigure}{0.475\linewidth}
		\includegraphics[width=1.\linewidth]{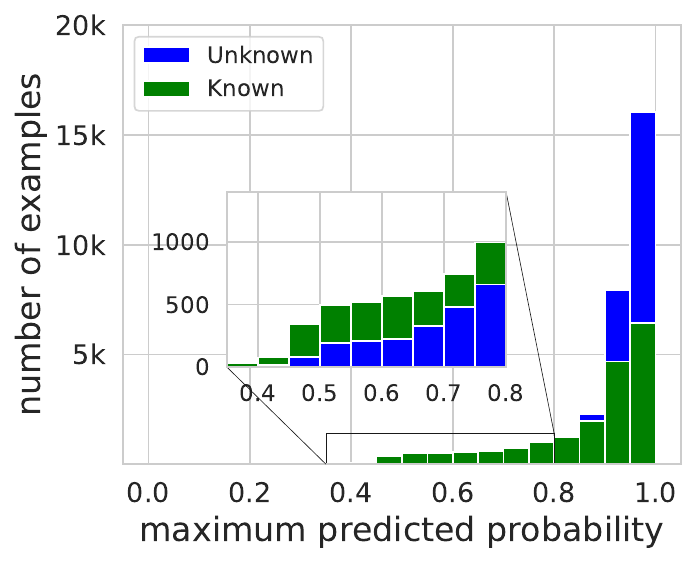}
		\caption{w/ NL}
		\label{fig.2b}
	\end{subfigure}
	\caption{The statistics of prediction confidence before and after fine-tuning the model. In the zoomed-in area of Figure \ref{fig.2b}, we swapped the display order of the two to prevent occlusion, allowing for a more intuitive view of how the distribution has changed.}
	\label{fig.2}
\end{figure}

To validate this, we conducted preliminary experiments on CIFAR-10~\cite{krizhevsky2009learning} with 4 known and 6 unknown classes. Figure \ref{fig.2} illustrates the confidence statistics for all unlabeled examples in a fixed query round before and after fine-tuning the model. As expected, known class examples are more prevalent in the low-confidence region, while unknown class ones are more common in high-confidence regions. This distribution characteristic presents a potential solution to the aforementioned question and offers a promising approach to AL for open-set scenarios.

Based on this, we propose a Bidirectional Uncertainty-based Active Learning framework (BUAL). On the one hand, we propose the Random Label Negative Learning (RLNL) method to fine-tune the model and leverage information from the unlabeled data pool. Specifically, we first train a model using all labeled data as a positive classifier. Then, negative learning is performed to fine-tune the model as a negative classifier by randomly assigning labels to data from the unlabeled pool in each training iteration. This effectively distinguishes known class examples with lower confidence predictions from unknown class ones. On the other hand, we propose a Bidirectional Uncertainty (BU) sampling strategy for active selection, which estimates prediction uncertainty from both positive and negative classifiers. By selecting examples with the highest uncertainty, we expect to identify the most informative instances from the known classes. Experiments are performed on multiple datasets with different unknown class ratios. The results demonstrate that BUAL can query more informative known class examples and that the model performance obtained by BUAL is substantially improved compared with existing state-of-the-art methods.

\section{Related Work}

Active learning (AL) is a prominent approach aimed at reducing label costs by selecting a batch of examples that are most valuable for model training. Existing AL methods can be broadly categorized into three groups based on sample selection strategies: uncertainty-based, representative-based, and hybrid strategies which combine both aspects. Uncertainty-based strategies focus on sampling informative instances to reduce model uncertainty. Typical methods include Least Confident Sampling~\cite{li2006confidence}, Margin-based Sampling~\cite{balcan2007margin}, and Entropy-based Sampling~\cite{holub2008entropy}, \etc. Representative-based strategies start from the sample distribution and aim to select representative instances that match the overall distribution. A typical method is Coreset~\cite{sener2017active}. Hybrid strategies combine uncertainty and representativeness by incorporating sample distribution information and the model's specific needs. Notable methods in this category include QUIRE~\cite{huang2010active} and BADGE~\cite{ash2019deep}, \etc.

Open-set annotation (OSA) involves active learning under open-set scenarios~\cite{du2021contrastive,ning2022active}. Existing methods primarily focus on selecting examples that are most likely to belong to known classes. For instance, CCAL~\cite{du2021contrastive} employs contrastive learning to extract semantic and distinctive features of examples, facilitating discrimination of known class examples. LfOSA~\cite{ning2022active} introduces an auxiliary network to model the per-example max activation value (MAV) distribution and dynamically selects examples with the highest probability from known classes. However, these methods exhibit sensitivity to the openness of the dataset and may not consistently perform well. Open-set recognition (OSR)~\cite{scheirer2012toward,salehi2021unified,moon2022difficulty} is a related problem setting to OSA, aiming to predict correct labels for known class examples while simultaneously detecting examples from unknown classes. Nevertheless, direct use of OSR methods often falls short of expected effectiveness~\cite{ning2022active}, mainly due to limited training examples and the inability to identify highly informative instances.

Complementary labels are labels other than the ground-truth label assigned to an example. Complementary label learning (CLL)~\cite{ishida2017learning,yu2018learning,feng2020learning} was initially introduced in~\cite{ishida2017learning}, where the authors leveraged the lower acquisition cost and higher correct labeling rate of complementary labels to address multi-class classification problems. However, complementary labels contain less information compared to the correct labels, which can result in slow convergence of the model and challenges in achieving the desired performance when directly using complementary labels for training. Authors in~\cite{kim2019nlnl} made the first attempt to combine CLL with the noise labeling problem~\cite{zong2024dirichlet} and proposed an indirect and robust training schema called negative learning (NL). In this paper, we exploit the properties of NL and extend it to open-set and label-free settings for the first time.

\section{Methodology}
\label{methods}

\begin{figure}[tb]
	\centering
	\includegraphics[width=0.99\textwidth]{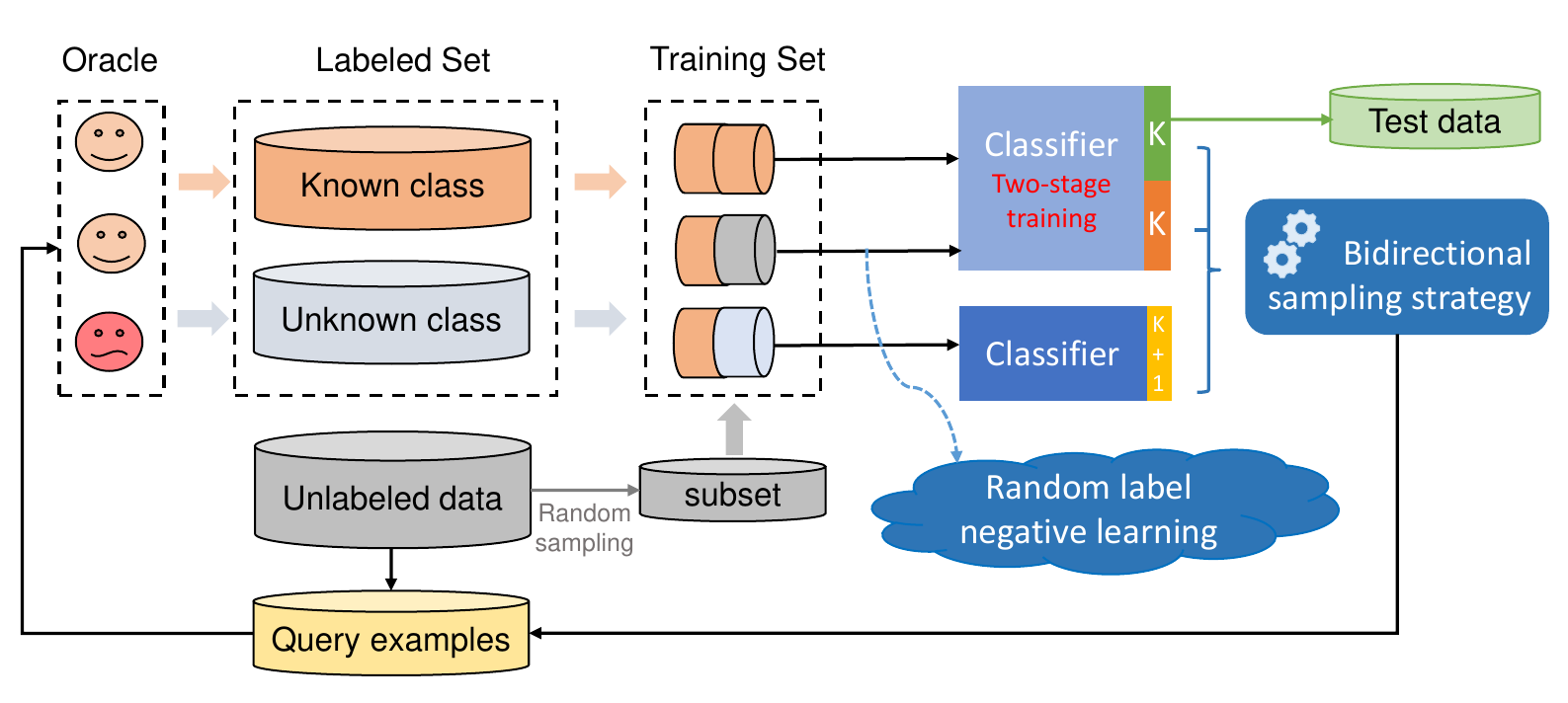}
	\caption{The framework of BUAL. A two-stage $K$ class classifier is maintained, where the first stage is trained in a normal manner saved as $f_p(\cdot)$ and the second stage is trained using the proposed random label negative learning method denoted as $f_n(\cdot)$. An auxiliary $K+1$ class classifier $f_{aux}(\cdot)$ is trained in parallel. By collecting the predicted uncertainty from $f_p(\cdot)$ and $f_n(\cdot)$ on each candidate example along with the global and local balancing factors, the proposed bidirectional sampling strategy can accurately estimate the potential utility of each example and perform effective sample sampling under complex open-set scenarios.}
	\label{fig.3}
\end{figure}

\subsection{Preliminaries}
\textbf{Notations.} In open-set annotation (OSA) settings, there are two labeled data pools: $D_{l}^{kno} =\left \{  (x_{i}^{l}, y_{i}^{l} ) \right \}_{i=1}^{n_{l}^{kno} }$ for known classes and $D_{l}^{unk} =\left \{  (x_{i}^{l}) \right \}_{i=1}^{n_{l}^{unk} }$ for unknown classes, as well as an unlabeled data pool: $D_{u} = \left \{  (x_{i}^{u} ) \right \}_{i = 1}^{n_{u}^{kno} } \cup  \left \{  (x_{i}^{u} ) \right \}_{i = 1}^{n_{u}^{unk} }$ containing examples from both known and unknown classes. Each known class example belongs to one of $K$ known classes in the label space $\mathcal{Y} =\left \{ 1,2,\dots, K \right \} $. The unknown class examples are uniformly grouped into one category, denoted as $y=\emptyset $. In each training round, active learning (AL) queries a batch of $b$ examples according to a given query strategy $\mathcal{A}$, denoted as $X_{query} = X_{query}^{kno} \cup X_{query}^{unk}$. Once the labeled feedback is obtained, we can calculate a ratio $r = \frac{|X_{query}^{kno}|}{|X_{query}|} $ to represent the precision of known classes.

\textbf{Overview.} The proposed Bidirectional Uncertainty-based Active Learning (BUAL) framework is depicted in Figure \ref{fig.3}. In each iteration, there are in general three steps: 
\begin{itemize}
\item Model Training: the algorithm first trains a target classifier in a normal learning manner, dubbed positive classifier $f_p(\cdot )$. Then, the algorithm fine-tunes the classifier with a new classifier head by the proposed \textit{Random Label Negative Learning} method, dubbed negative classifier $f_n(\cdot )$. Similar to~\cite{ning2022active}, a $K+1$ auxiliary classifier $f_{aux}(\cdot )$ is trained in parallel.

\item Example Selection: the \textit{Bidirectional Uncertainty Sampling Strategy} estimates uncertainty bidirectionally using both the positive and negative classifier heads. This estimation is combined with dynamic balance factors generated by the auxiliary classifier and query feedback to select the most informative known class examples.

\item Oracle Labeling: the annotators assign class labels to the selected examples. Based on the feedback results, update the corresponding data pools accordingly.
\end{itemize}

\subsection{Random Label Negative Learning}

In OSA scenarios, conventional uncertainty-based AL methods tend to be ineffective for example selection, mainly due to the unconfident predictions generated for the unknown class examples. Existing OSA methods prioritize sample purity without fully exploring sample informativeness, resulting in queries with too many already mastered examples that are not useful for model training. Thus, a key question arises: how to distinguish between unknown and "informative" known class examples.

To cope with this problem, we propose a general method by pushing unknown class examples toward the high-confidence regions, while pushing known class examples to the low-confidence regions. If this separation is achieved, we can leverage existing uncertainty-based AL methods directly to handle various complex open-set scenarios. Fortunately, we achieve this to some extent by employing negative learning (NL), where we assign random labels to unlabeled examples and fine-tune the model with a new classifier head accordingly, dubbed Random Label Negative Learning (RLNL).

Specifically, in the model training stage, we first train a $K$-class classifier $f_p(\cdot )$ in a normal training manner (\eg, cross-entropy loss) based on $D_l^{kno}$. Note that $f_p(\cdot )$ can be the target model we eventually need to output. After training on the labeled data, $f_p(\cdot)$ can own a good discriminative ability for known class examples at the representation level. Then, we replace the last layer of $f_p(\cdot)$ and fine-tune a new classifier head $f_n(\cdot)$ with Equation \ref{eq1}. In this phase, unlabeled examples in $D_u$ are involved in training to help achieve the goal mentioned above: pushing unknown and known class examples toward the high and low confidence regions, respectively. All labeled examples in $D_l^{kno}$ are still involved to prevent them from being incorrectly shifted in the distribution. Eventually, we assign random labels to the unlabeled examples in $D_u$, while using the complementary labels instead for the labeled known class examples, \ie, for an example $\boldsymbol{x}$:
\begin{equation}\label{eq2}
P(\bar{y}=s)=\frac{1}{\left | S \right | }, s  \in \begin{cases}
 S= \mathcal{Y}\setminus y^l, & \text{ if } \boldsymbol{x}\in D_{l}^{kno}, \\
 S= \mathcal{Y}, & \text{ if } \boldsymbol{x}\in D_u.
\end{cases}
\end{equation}
where $\bar{y}$ is uniformly sampled at each training iteration. Here, considering that the number of examples in $D_u$ is usually much larger than that in $D_{l}^{kno}$, a small subset $D_{sub}$ is randomly selected from the $D_u$ to save training cost.

\begin{figure}[tb]
	\centering
	\includegraphics[width=0.85\linewidth]{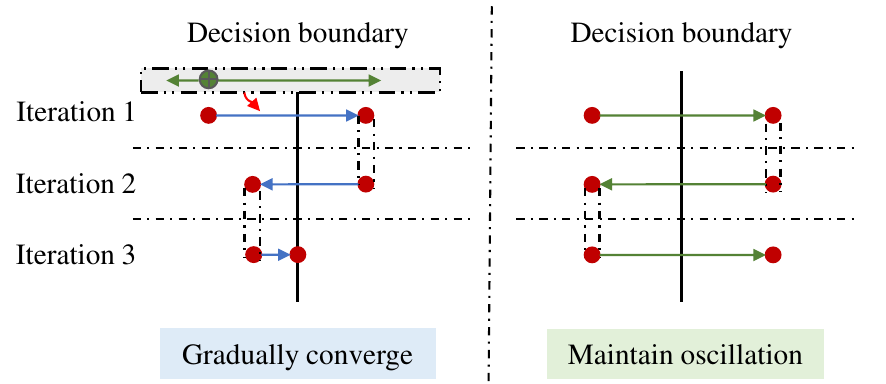}
	\caption{Use all labels per iteration for negative learning (left) vs. use one random label per iteration for negative learning (right).}
	\label{fig.4}
\end{figure}

\textbf{Why RLNL works?} The first question one may hold is why it shouldn't be optimal for the unlabeled examples to be the uniform distribution as $\bar{y}$ is sampled at each iteration. To explain this, we take two different kinds of updating for a single example in the binary classification scenario of "Dog vs. Cat" as illustrations: negative learning using all labels at each iteration and negative learning using only one random label at each iteration. One possible case is shown in Figure \ref{fig.4}. If each iteration uses all labels for negative learning, \ie, such an example is neither a "Cat" nor a "Dog", it is apparent that an update gradient of 0 is only achieved when the example is at the decision boundary. In contrast, if each iteration gives only one random label for negative learning, \ie, such an example is not a "Cat" or a "Dog", then for an instance that is not a "Cat", pushing it as much as possible towards the "Dog" side will result in a gradient update of 0, and vice versa. Obviously, there is no fixed optimal scenario for this type of data.

In RLNL, we utilize this property by exploiting the prior knowledge contained in the labeled data. For unlabeled known class data, they might have overlapping regions with labeled data in the feature space, benefiting from the feature representations learned in the previous stage and the simultaneous introduction of labeled data for negative learning. The mapping of such examples in the feature space will remain or be close to the labeled ones in subsequent model updates, as they will receive invisible constraints provided by the prior knowledge from the labeled data, owing to similar features.

\begin{figure}[tb]
	\centering
	\includegraphics[width=0.7\linewidth]{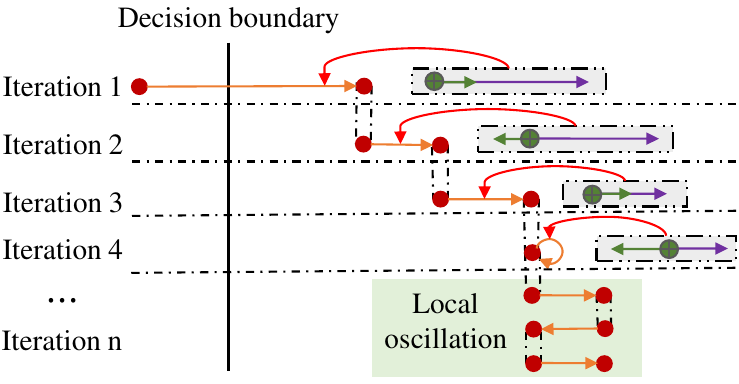}
	\caption{The possible RLNL update scenario for unlabeled unknown class data in batch deep learning manner. The green "$\longrightarrow $" is the batch update gradient produced by the example itself, and the purple "$\longrightarrow $" is the update gradient produced by labeled data. Initially, the decision boundary is close to the left-hand category.}
	\label{fig.5}
\end{figure}

\begin{figure}[tb]
	\centering
	\begin{subfigure}{0.45\linewidth}
		\includegraphics[width=1.\linewidth]{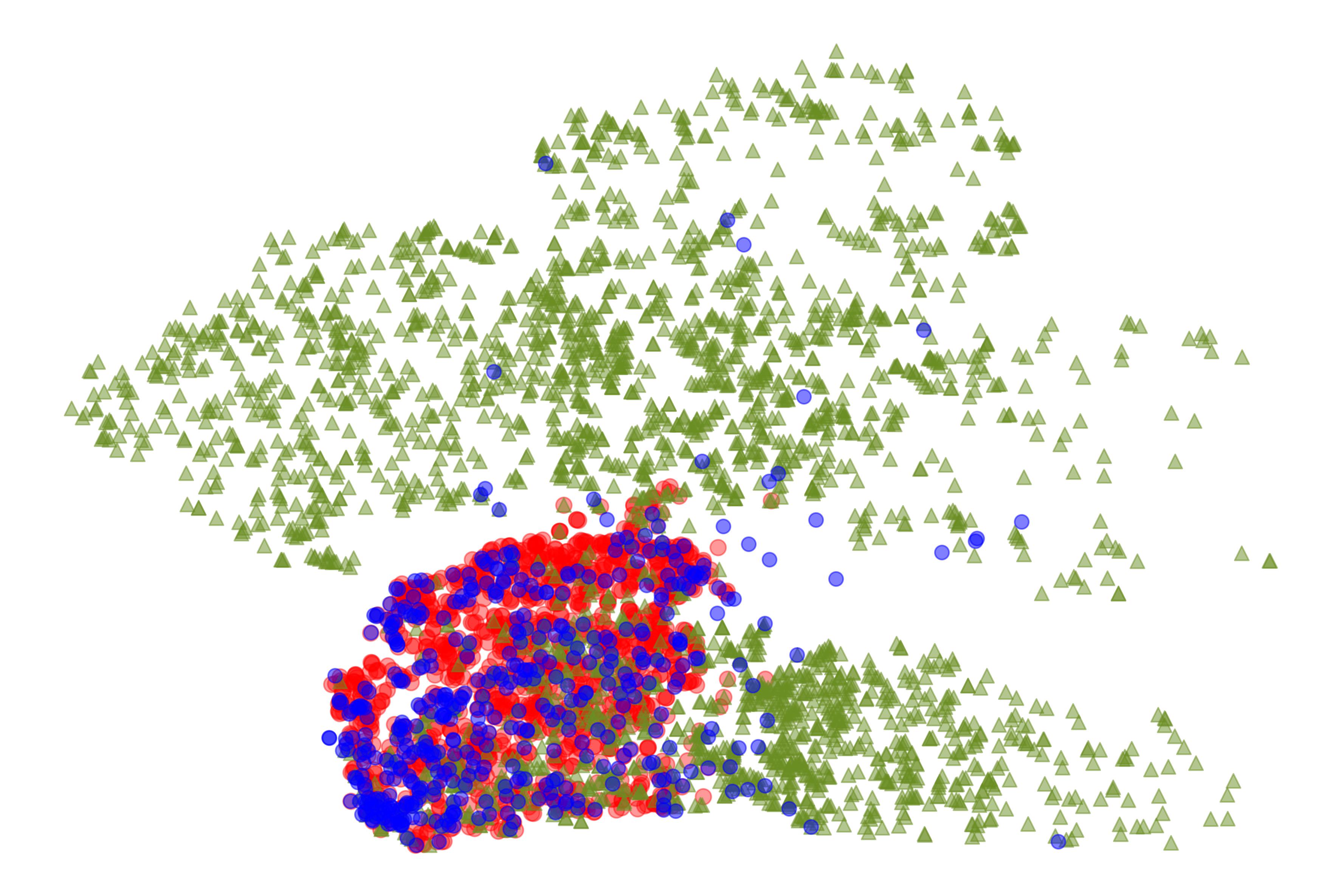}
		\caption{Before RLNL}
		\label{fig.tsne2_a}
	\end{subfigure}
	\hfill
	\begin{subfigure}{0.54\linewidth}
		\includegraphics[width=1.\linewidth]{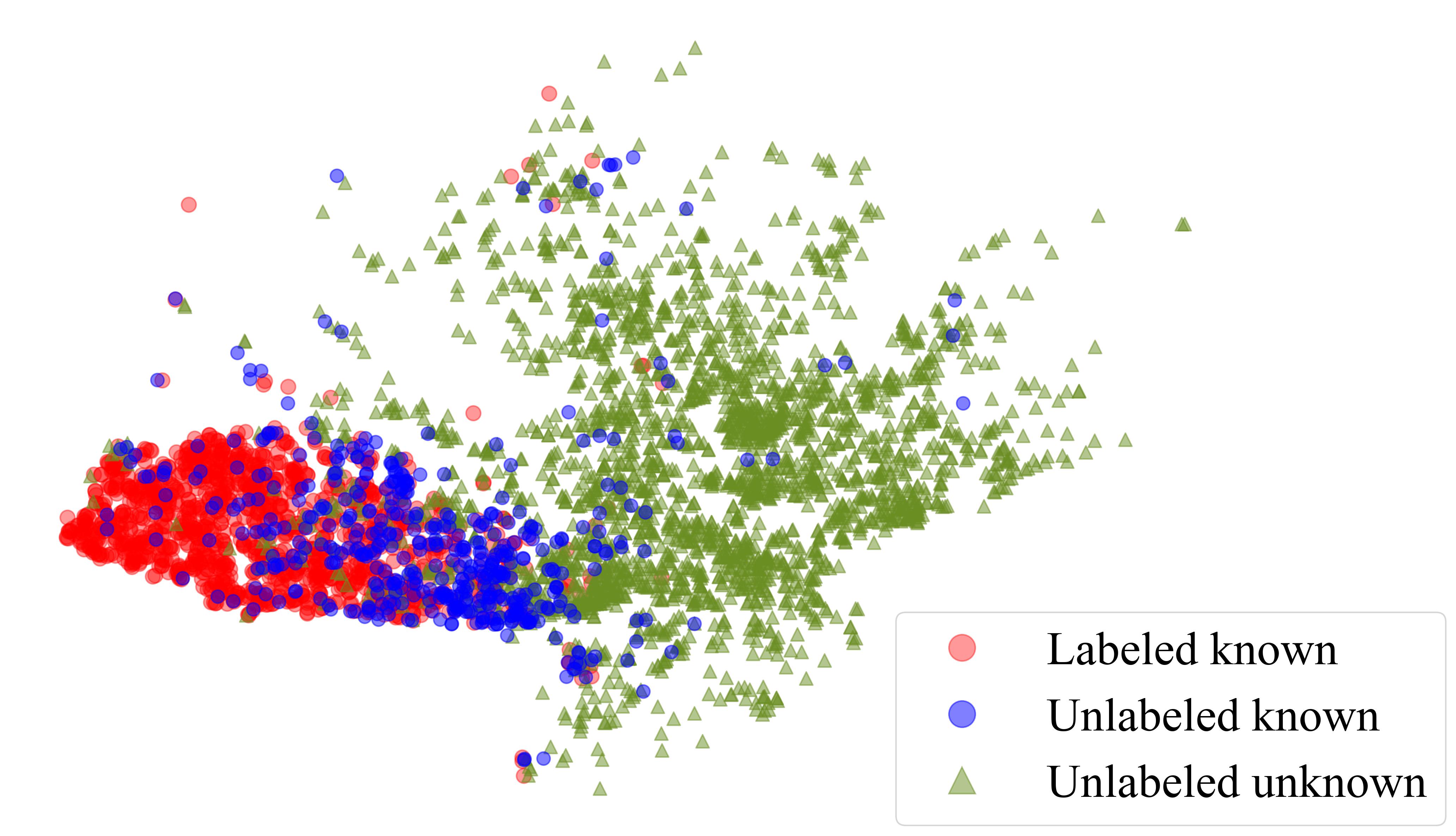}
		\caption{After RLNL}
		\label{fig.tsne2_b}
	\end{subfigure}
	\caption{The t-SNE feature visualization of labeled data, unlabeled known class data, and unlabeled unknown class data on CIFAR-10 with an openness ratio of 0.5 before and after performing RLNL. For a more intuitive visualization, we only show a single known class. More visualization results are shown in the supplementary file.}
	\label{fig.tsne2}
\end{figure}

As shown in Figure \ref{fig.5}, we present the possible update scenario for unlabeled unknown class examples in the commonly adopted batch deep learning manner. Here, the green arrow indicates the batch update gradient produced by itself, while the purple arrow indicates the batch update gradient produced by labeled data. Different from Figure \ref{fig.4}, where data oscillates on both sides of the decision boundary, the unlabeled unknown class examples will oscillate at uncharted away from the decision boundary to counteract the update gradient due to the labeled ones. In contrast, unlabeled known class examples will move much less than unknown class ones in magnitude in the feature space within the constraints of the prior knowledge provided by the labeled data. This is further confirmed in Figure \ref{fig.tsne2}, which illustrates the t-SNE feature visualization of labeled data, unlabeled known class data, and unlabeled unknown class data before and after performing RLNL. This ultimately leads to the result in Figure \ref{fig.2b} and proves that RLNL does work.

\subsection{Bidirectional Uncertainty Sampling Strategy}

During the fine-tuning process, we observed that the predictions of $f_n(\cdot )$ for examples in $D_u$ tend to oscillate between epochs. To ensure stable sampling, on one hand, we test all unlabeled examples $t$ times at $m$ round intervals to obtain the predicted probabilities $\boldsymbol{p_t^{-}} = (p_1^t, ..., p_K^t)$, which will further be averaged as $\boldsymbol{\mathcal{P}^{-}} = \frac{1}{t} \sum_{i=1}^{t} \boldsymbol{p_i^{-}}=(\frac{1}{t} \sum_{i=1}^{t}p_1^i,...,\frac{1}{t} \sum_{i=1}^{t}p_K^i)$. On the other hand, we reuse the predictions of $f_p(\cdot)$ to produce predicted probabilities $\boldsymbol{p^{+}} = (p_1, ..., p_K)$ for all unlabeled examples, as $f_p(\cdot)$ are accurate for measuring known class ones. 

Compared to the positive head, the negative head is slightly biased for the measurement of sample uncertainty due to the unstable training. Therefore, if an example is likelier to belong to known classes, we prefer to utilize the sample uncertainty obtained from $f_p(\cdot)$. On the contrary, once an example has a higher risk of belonging to the unknown classes, the uncertainty obtained from $f_p(\cdot)$ is unreliable, and thus the uncertainty by $f_n(\cdot)$ should be given a higher weight. 

To achieve this, we introduce two balancing factors based on global and local observations. In open-set scenarios, some unknown class examples might be mistakenly selected for annotation. Although these examples cannot be directly used for training the target model, they are valuable for measuring sample purity. Therefore, similar to~\cite{ning2022active}, we train a $K+1$ classifier $f_{aux}(\cdot)$ in a normal training manner based on the examples from both $D_{l}^{kno}$ and $D_{l}^{unk}$. Then, we can obtain the predicted probability $p_{K+1}^{aux}(\boldsymbol{x}) = f_{aux}(\emptyset  \mid \boldsymbol{x})$ for each example. This can act as a local balancing factor, since the larger the value of $p_{K+1}^{aux}(\boldsymbol{x})$ the more likely $\boldsymbol{x}$ is to belong to the unknown class.

Additionally, once the selected examples are sent to the oracle for annotation, we can calculate the ratio $r$, which provides a rough estimate of the current openness of $D_u$ and can serve as a global balancing factor. With the two balancing factors, we propose a bidirectional uncertainty sampling strategy defined as follows:
\begin{equation}\label{eq11}
	\boldsymbol{x}^*=arg\max_{\boldsymbol{x}} p_{K+1}^{aux}(\boldsymbol{x})unc_{n}+r\left [1-p_{K+1}^{aux}(\boldsymbol{x})\right ]unc_{p} , 
\end{equation}
where $unc_{p}$ and $unc_{n}$ denote the uncertainty of $\boldsymbol{x}$ to the positive classifier head and the negative classifier head, respectively. Noteworthy, this sampling strategy remains applicable even in closed-set settings. If there are no unknown class examples, the ratio $r$ will always be equal to 1, and $p_{K+1}^{aux}(\boldsymbol{x})$ will be 0, effectively making the strategy equivalent to a normal uncertainty sampling strategy.

With Equation \ref{eq11}, extending the existing closed-set uncertainty-based active learning methods to open-set scenarios is possible. In this paper, we focus on three classical uncertainty sampling strategies: least confident sampling, margin-based sampling, and entropy-based sampling. The corresponding modified versions for open-set scenarios are as follows:

\begin{itemize}
    \item[$\bullet$]\textbf{Bidirectional Least Confident Sampling}
\end{itemize}
\begin{equation}\label{eq3}
	\begin{aligned}
		\boldsymbol{x}^*= \mathop{\arg\max}_{\boldsymbol{x}}p_{K+1}^{aux}(\boldsymbol{x})\left [  1-\boldsymbol{\mathcal{P}}_{y^{-}}^{-}(\boldsymbol{x} ) \right ]  +r\left [ 1-p_{K+1}^{aux}(\boldsymbol{x})\right ]\left [  1- \boldsymbol{p}_{y^{+}}^{+}( \boldsymbol{x} ) \right ] ,
\end{aligned}
\end{equation}
where $y^{-}=\mathop{\arg\max}_{y}\boldsymbol{\mathcal{P}}_{y}^{-}(\boldsymbol{x} ) $, $y^{+}=\mathop{\arg\max}_{y} \boldsymbol{p}_{y}^{+}(\boldsymbol{x} ) $.

\begin{itemize}
    \item[$\bullet$]\textbf{Bidirectional Margin-Based Sampling}
\end{itemize}
\begin{equation}\label{eq4}
	\begin{aligned}
		\boldsymbol{x}^*= & \mathop{\arg\max}_{\boldsymbol{x}}p_{K+1}^{aux}(\boldsymbol{x})\left [ \boldsymbol{\mathcal{P}}_{y_1^{-}}^{-}( \boldsymbol{x} )-\boldsymbol{\mathcal{P}}_{y_2^{-}}^{-}( \boldsymbol{x} ) \right ]  \\& +r\left [1-p_{K+1}^{aux}(\boldsymbol{x})\right ]\left [ \boldsymbol{p}_{y_1^{+}}^{+}(\boldsymbol{x} )- \boldsymbol{p}_{y_2^{+}}^{+}(\boldsymbol{x} ) \right ] ,
\end{aligned}
\end{equation}
where $y_1^{-}=\mathop{\arg\max}_{y}\boldsymbol{\mathcal{P}}_{y}^{-}(\boldsymbol{x}) $, $y_1^{+}=\mathop{\arg\max}_{y}\boldsymbol{p}_{y}^{+}(\boldsymbol{x} ) $ $y_2^{-}=\mathop{\arg\max}_{y \setminus y_1^{-}}\boldsymbol{\mathcal{P}}_{y}^{-}(\boldsymbol{x}) $, $y_2^{+}=\mathop{\arg\max}_{y \setminus y_1^{+}} \boldsymbol{p}_{y}^{+}(\boldsymbol{x} ) $.

 \begin{itemize}
    \item[$\bullet$]\textbf{Bidirectional Entropy-Based Sampling}
\end{itemize}
\begin{equation}\label{eq5}
	\begin{aligned}
		\boldsymbol{x}^*= & \mathop{\arg\max}_{\boldsymbol{x}}p_{K+1}^{aux}(\boldsymbol{x})\left [-\boldsymbol{\mathcal{P}}_{y^{-}}^{-}(\boldsymbol{x} )\log_{}{\boldsymbol{\mathcal{P}}_{y^{-}}^{-}(\boldsymbol{x} )} \right ] \\ & +r\left [1-p_{K+1}^{aux}(\boldsymbol{x})\right ]\left [ -\boldsymbol{p}_{y^{+}}^{+}( \boldsymbol{x} )\log_{}{ \boldsymbol{p}_{y^{+}}^{+}( \boldsymbol{x} )} \right ] ,
	\end{aligned}
\end{equation}
where $y^{-}$ and $y^{+}$ are consistent with the previous definition.

The main procedures of BUAL are summarized in Algorithm \ref{alg:2}.

\begin{algorithm}[tb]
	\caption{BUAL Training Procedure}
	\label{alg:2}
	\textbf{Input}: $D_l^{kno}$, $D_l^{unk}$, $D_u$, $r$, $\mathcal{A}$, $b$, $m$, $t$, subset size $s$, training epoch $E$, optimizer $\mathcal{O} $.  \\
	\textbf{Output}: $D_l^{kno}$, $D_l^{unk}$, $D_u$, $r$, model parameters $\theta_p$.\\
	\textbf{Process:}
	\begin{algorithmic}[1]
		\STATE \textbf{\emph{\# Model training}}
		\FOR {$j=1;j\le E$}
		\STATE $\mathcal{L}_p= {\textstyle \sum_{(x,y)\in D_l^{kno}}^{}\ell(f_p(\boldsymbol{x}),y)}  $;
		\STATE $\mathcal{L}_{aux}= {\textstyle \sum_{(x,y)\in D_l^{kno}\cup D_l^{unk}}^{}\ell(f_{aux}(\boldsymbol{x}),y)}$;
		\STATE Update $\theta_p, \theta_{aux} = \mathcal{O}(\mathcal{L}_p, \theta_p), \mathcal{O}(\mathcal{L}_{aux}, \theta_{aux})$;
		\ENDFOR
		\STATE Use $f_{aux}(\boldsymbol{x})$ to remove confidently unknown examples in $D_u$ and randomly select $s$ examples as $D_{sub}$;
		\FOR {$j=1;j\le E$}
        \STATE Generate label $\bar{y}$ for each example.
		\STATE $\mathcal{L}_{n}= {\textstyle \sum_{(x,y)\in D_l^{kno}\cup D_{sub}}^{}\ell _{NL}(f_{n}(\boldsymbol{x}),\bar{y} )}$;
		\STATE Update $\theta_n = \mathcal{O}(\mathcal{L}_n, \theta_n)$;
		\IF{ $j$ mod $m = 0$ }
		\STATE Obtain $\boldsymbol{p}_j^{-}$ for each sample in $D_u$ by $f_n(\boldsymbol{x})$;
		\ENDIF
		\ENDFOR
		\STATE \textbf{\emph{\# Example selection}}
		\STATE Calculate $\boldsymbol{\mathcal{P}}^{-}$ for each sample across rounds;
		\STATE Obtain $\boldsymbol{p}^{+}$ for each sample in $D_u$ by $f_p(\boldsymbol{x})$;
		\STATE Obtain $p_{K+1}^{aux}$ for each sample in $D_u$ by $f_{aux}(\boldsymbol{x})$;
		\STATE Obtain $X_{query}$ using BU sampling strategy $\mathcal{A}$;
		\STATE \textbf{\emph{\# Oracle labeling}}
		\STATE Ask for annotation and update $D_l^{kno}$, $D_l^{unk}$, $D_u$ and $r$.
	\end{algorithmic}
\end{algorithm}

\section{Experiments}

\textbf{Datasets.} We conduct experiments on three benchmark datasets: CIFAR-10, CIFAR-100~\cite{krizhevsky2009learning}, and Tiny-Imagenet~\cite{yao2015tiny}. Tiny-Imagenet is a subset of the Imagenet~\cite{krizhevsky2012imagenet}, consisting of 200 classes with 500 training images per class. The openness of each dataset is defined as the ratio of unknown classes to the total number of classes, and we set its value to 0.2, 0.4, 0.6, and 0.8 for all datasets.

\begin{figure}[tb]
	\centering
	\includegraphics[width=5.25cm]{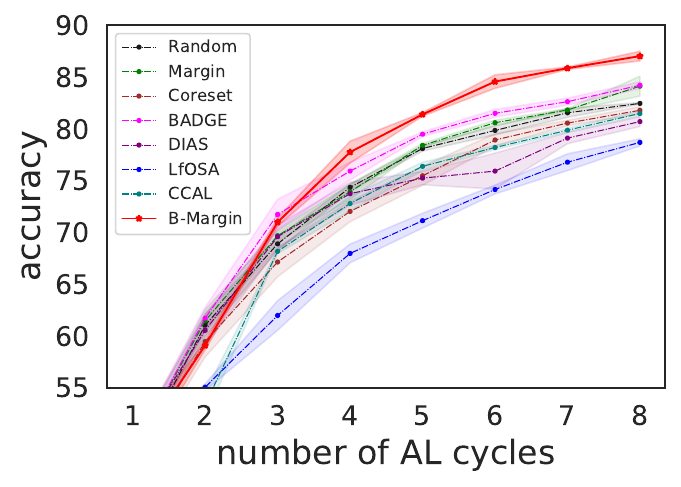}
        \includegraphics[width=5.25cm]{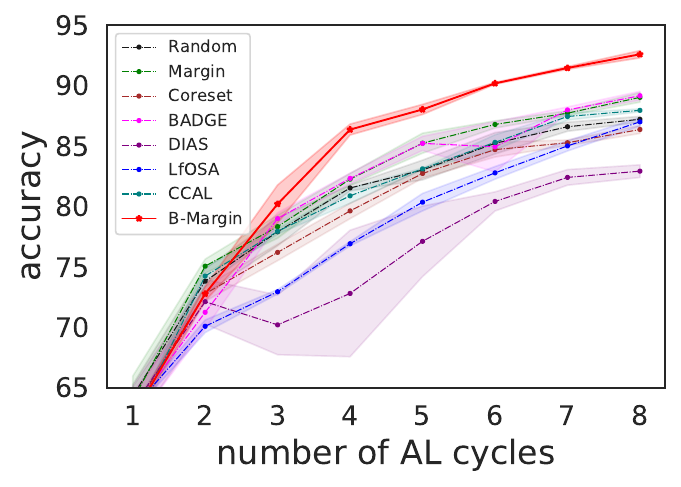}
        
        \includegraphics[width=5.25cm]{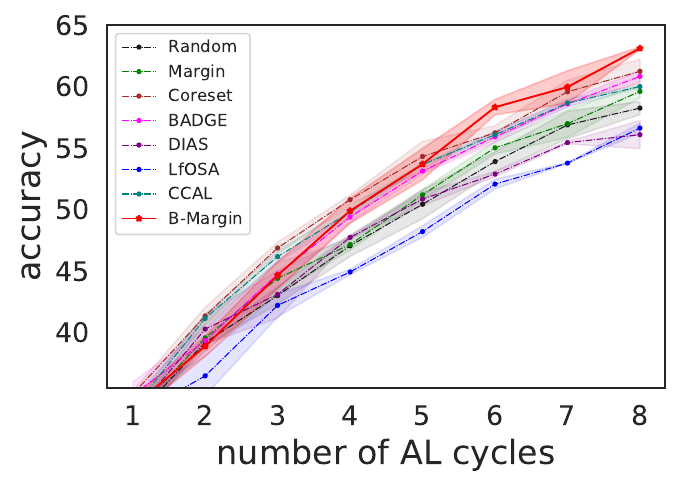}
        \includegraphics[width=5.25cm]{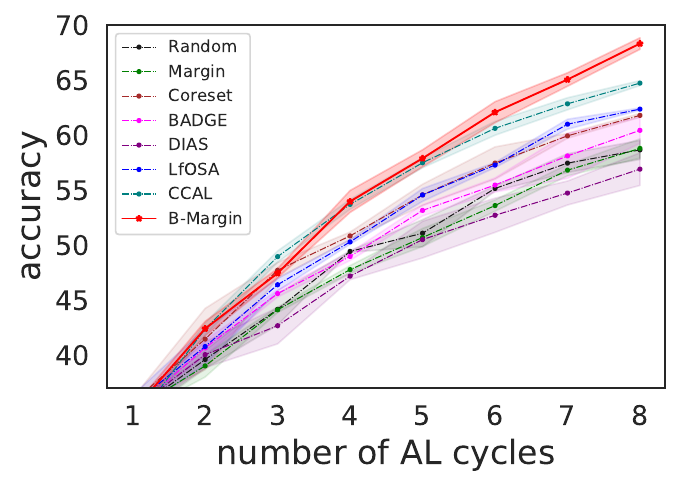}
        
        \includegraphics[width=5.25cm]{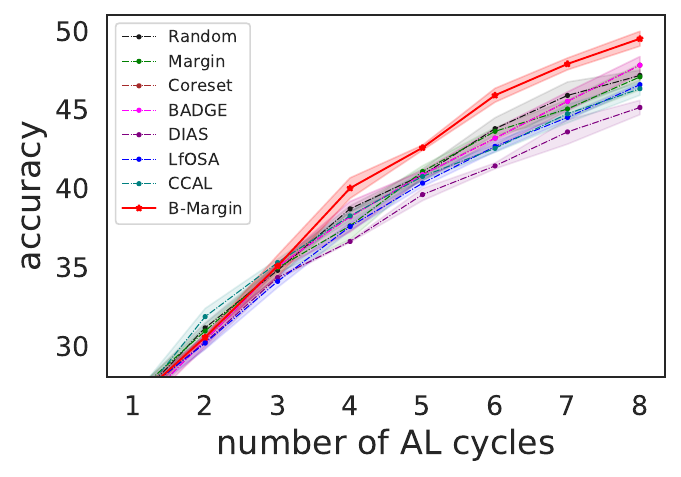}
	\includegraphics[width=5.25cm]{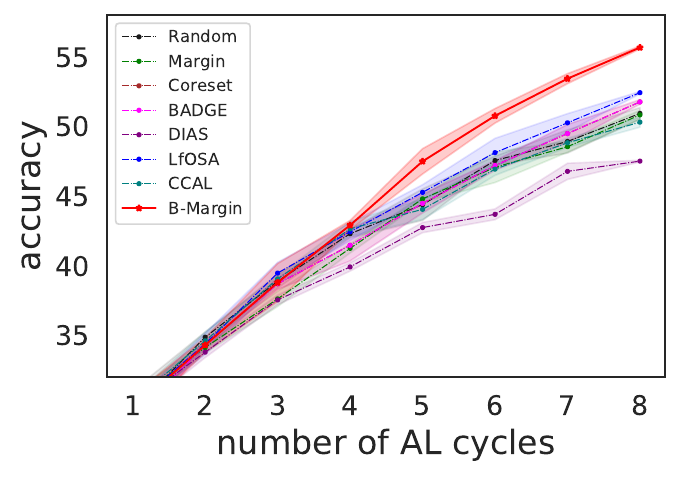}
	\caption{Accuracy comparison on CIFAR-10 (first row), CIFAR-100 (second row), and Tiny-Imagenet (third row). The ratio of unknown class examples to the total number of examples is fixed at 0.4 (first column) and 0.6 (second column) for each dataset. }
	\label{fig.acc}
\end{figure}

\textbf{Comparing methods.} We select nine AL strategies for comparison, which can be further categorized into six groups: (1) Random: Randomly select examples from the unlabeled data pool for labeling. (2) Traditional uncertainty-based strategies: Least confident sampling (LC), Margin-based sampling (Margin), and Entropy-based sampling (Entropy). (3) Diversity-based strategy: Coreset. (4) Hybrid-based strategy: BADGE. (5) OSA methods: CCAL and LfOSA. (6) OSR method: DIAS~\cite{moon2022difficulty}. Correspondingly, our methods are Bidirectional Least confident sampling (B-LC), Bidirectional Margin-based sampling (B-Margin), and Bidirectional Entropy-based sampling (B-Entropy).

\textbf{Training details.} On CIFAR-10, CIFAR-100, and Tiny-Imagenet, we randomly sample 1\%, 8\%, and 8\% known class examples as the initial labeled data, respectively. All the models involved in the experiments are ResNet18~\cite{he2016deep}, trained for 100 epochs, using SGD as the optimizer, where the learning rate is 0.01, momentum is 0.9, weight decay is 1e-4, and batch size is 128. We perform the experiments for 3 runs and report the average results. For CIFAR-10 and CIFAR-100, 5000 examples are randomly selected from the unlabeled pool as $D_{sub}$, and 1500 examples are queried in each query round. Due to the doubling of data volume, we randomly select 10000 examples as $D_{sub}$ and query 3000 examples in each query round for Tiny-Imagenet. 

\subsection{Performance Comparison}

Figure \ref{fig.acc} presents the variation curves for the classification accuracy of the proposed methods and the comparison methods. Here, to better observe the variation curves, we only show the B-Margin for the proposed methods and the Margin for the compared traditional uncertainty methods. Table \ref{table:acc} reports all methods' final round average accuracy.

\begin{table}[!t]	
	\centering
 \caption{The final round average accuracy of different methods on CIFAR-10, CIFAR-100, and Tiny-Imagenet. The best performance is highlighted in bold.}
	\label{table:acc}
 \addtolength{\tabcolsep}{2pt}
        \small
	\resizebox{0.99\columnwidth}{!}{
 \begin{tabular}{c|c|c c c c|c c c c|c c c c}\toprule \toprule
 \multicolumn{2}{c|}{Datasets}&\multicolumn{4}{c|}{CIFAR-10}&\multicolumn{4}{c|}{CIFAR-100}&\multicolumn{4}{c}{Tiny-Imagenet}\\
 \midrule
    \multicolumn{2}{c|}{Openness Ratio} &0.2&0.4&0.6&0.8&0.2&0.4&0.6&0.8&0.2&0.4&0.6&0.8\\ 	
    \midrule
    (1)&Random&$83.3$&$82.5$&$87.2$&$96.9$&$57.6$&$58.3$&$58.7$&$61.2$&$45.7$&$47.2$&$50.9$&$55.0$\\
 \midrule

  \multirow{3}{*}{(2)}&LC&$84.3$&$81.6$&$87.5$&$96.2$&$55.8$&$54.6$&$54.0$&$56.2$&$44.8$&$45.9$&$48.4$&$51.6$\\
&Margin&$86.0$&$84.1$&$89.0$&$97.0$&$59.3$&$59.6$&$58.8$&$58.9$&$46.4$&$47.1$&$50.8$&$54.0$\\
&Entropy&$85.4$&$83.4$&$88.0$&$96.8$&$57.1$&$56.8$&$55.7$&$56.4$&$44.6$&$44.5$&$46.9$&$50.7$\\
 \midrule

(3)&Coreset&$85.0$&$81.8$&$86.4$&$97.4$&$60.2$&$61.2$&$61.8$&$64.2$&$46.2$&$47.8$&$51.8$&$54.0$\\
 \midrule

(4)&BADGE&$86.8$&$84.2$&$89.2$&$96.4$&$60.2$&$60.8$&$60.4$&$62.0$&$46.3$&$47.8$&$51.8$&$53.3$\\
 \midrule

 \multirow{2}{*}{(5)}&LfOSA&$73.7$&$78.7$&$87.0$&$98.6$&$52.3$&$56.6$&$62.4$&$68.2$&$42.5$&$46.6$&$52.4$&$59.9$\\
&CCAL&$80.8$&$81.5$&$88.0$&$98.1$&$55.9$&$60.0$&$64.7$&$67.7$&$44.4$&$46.3$&$50.3$&$57.0$\\
 \midrule
 
 (6)&DIAS&$81.8$&$80.7$&$83.0$&$94.0$&$55.7$&$56.1$&$56.9$&$57.2$&$43.1$&$45.1$&$47.5$&$54.4$\\
 \midrule
 
\multirow{3}{*}{Ours}&B-LC&$\mathbf {87.0}$&$87.2$&$92.5$&$\mathbf{99.1}$&$59.3$&$62.8$&$67.5$&$\mathbf{72.1}$&$45.7$&$48.7$&$54.7$&$60.6$\\
 &B-Margin&$86.5$&$87.0$&$\mathbf{92.6}$&$98.9$&$\mathbf{60.9}$&$\mathbf{63.1}$&$\mathbf{68.3}$&$71.5$&$\mathbf{46.5}$&$\mathbf{49.5}$&$\mathbf{55.7}$&$\mathbf{61.2}$\\
 &B-Entropy&$86.9$&$\mathbf{87.4}$&$\mathbf{92.6}$&$\mathbf{99.1}$&$58.9$&$61.7$&$66.9$&$71.4$&$45.4$&$47.5$&$55.2$&$61.0$\\
    \bottomrule \bottomrule
	\end{tabular}}
\end{table}

\begin{figure}[tb]
	\centering
	
	\includegraphics[width=3.9cm]{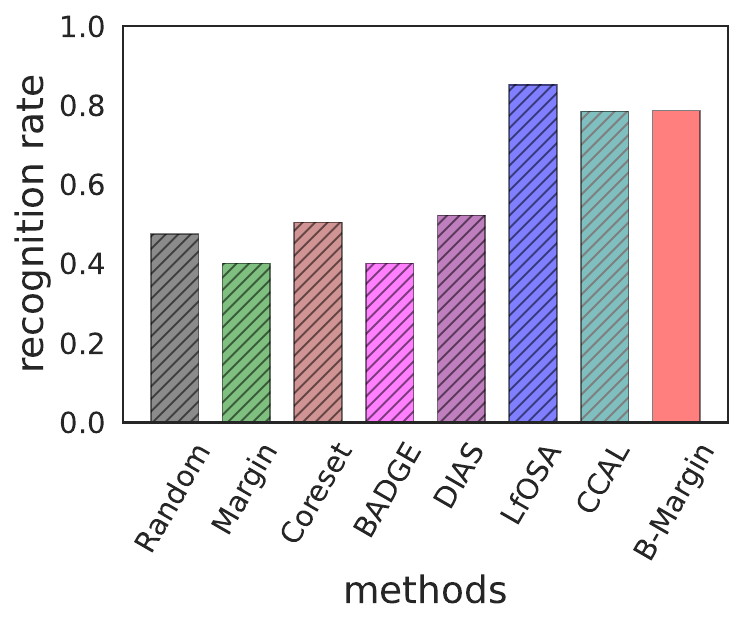}
	\includegraphics[width=3.9cm]{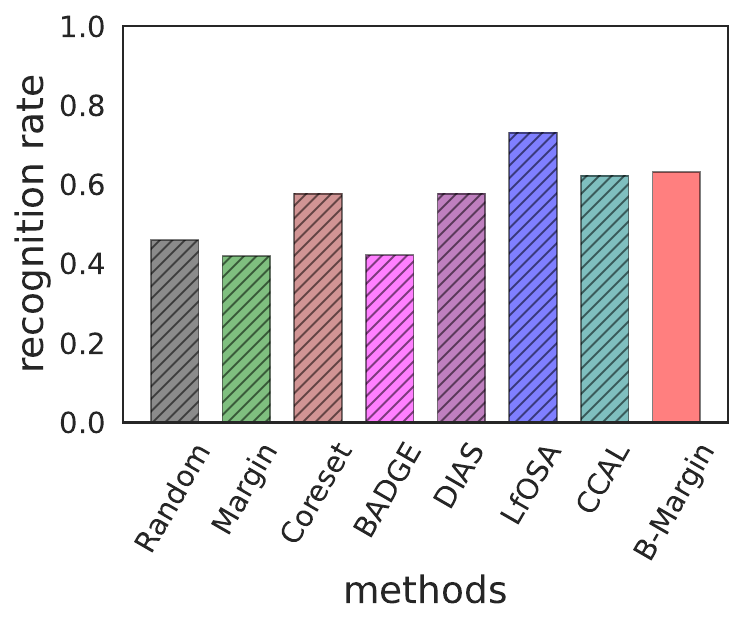}
	\includegraphics[width=3.9cm]{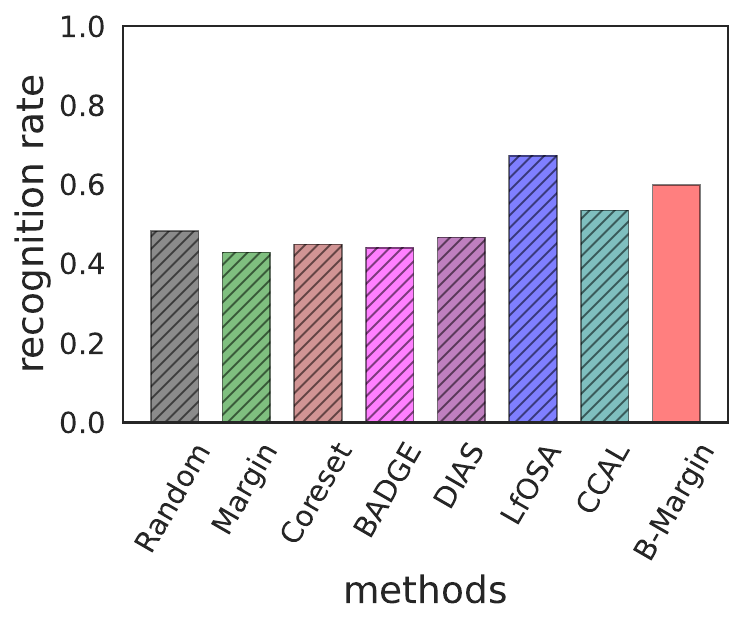}
	\caption{The average recognition rate on CIFAR-10 (first column), CIFAR-100 (second column), and Tiny-Imagenet (third column).}
	\label{fig.querys}
\end{figure}

We can observe that our proposed methods consistently achieve the highest classification accuracy, demonstrating the effectiveness and superiority of our BUAL framework over other methods. Some notable observations are as follows: 1) The OSA methods, CCAL and LfOSA, gradually lose effectiveness as the proportion of known classes in the dataset increases. In contrast, traditional uncertainty-based methods perform better as the openness ratio decreases. This is consistent with our previous analysis. 2) The performance of Coreset and BADGE also deteriorates when the openness ratio is high. These methods tend to select examples with diverse characteristics. However, unknown class examples often differ significantly in characteristics from the labeled ones, making them more likely to be sampled and thus undermining the effectiveness of this type of method. 3) The OSR method DIAS does not perform well in all situations. The limited labeled data prevents the model from learning a robust representation for identifying open-set examples effectively. Moreover, DIAS cannot identify highly informative examples, resulting in its queries often being simple and unhelpful examples. 4) All our methods remain stable and do not suffer significantly from changes in the openness ratio. This can demonstrate two points. The dynamic balance factors in the proposed framework can adaptively assign appropriate weights to positive and negative uncertainties regardless of the openness of the dataset, and negative uncertainties are indeed effective for querying highly informative known class examples.

\begin{figure}[tb]
	\centering
	\begin{subfigure}{0.45\linewidth}
		\includegraphics[width=1.\linewidth]{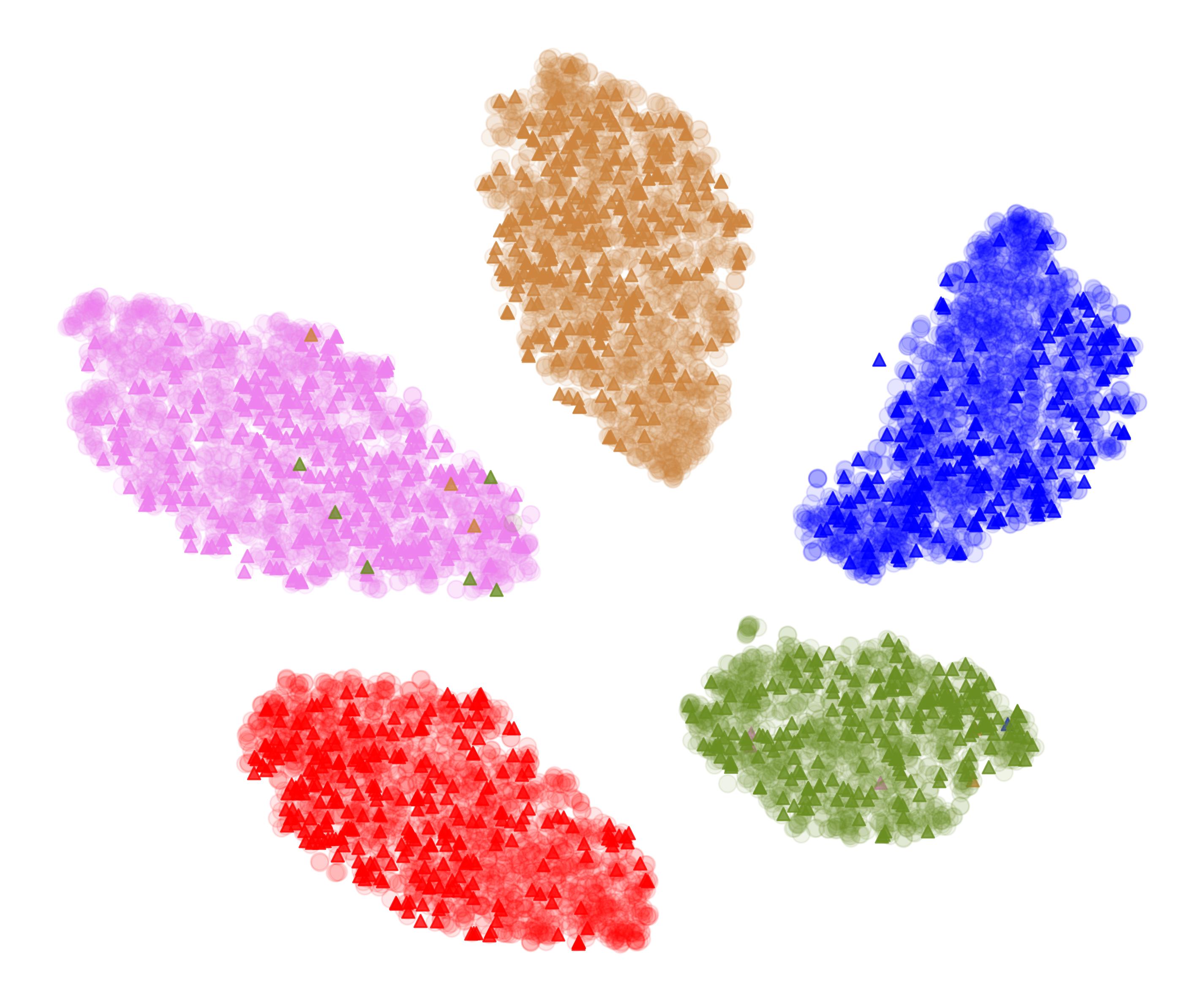}
		\caption{LfOSA}
		\label{fig.tsne_a}
	\end{subfigure}
	\begin{subfigure}{0.45\linewidth}
		\includegraphics[width=1.\linewidth]{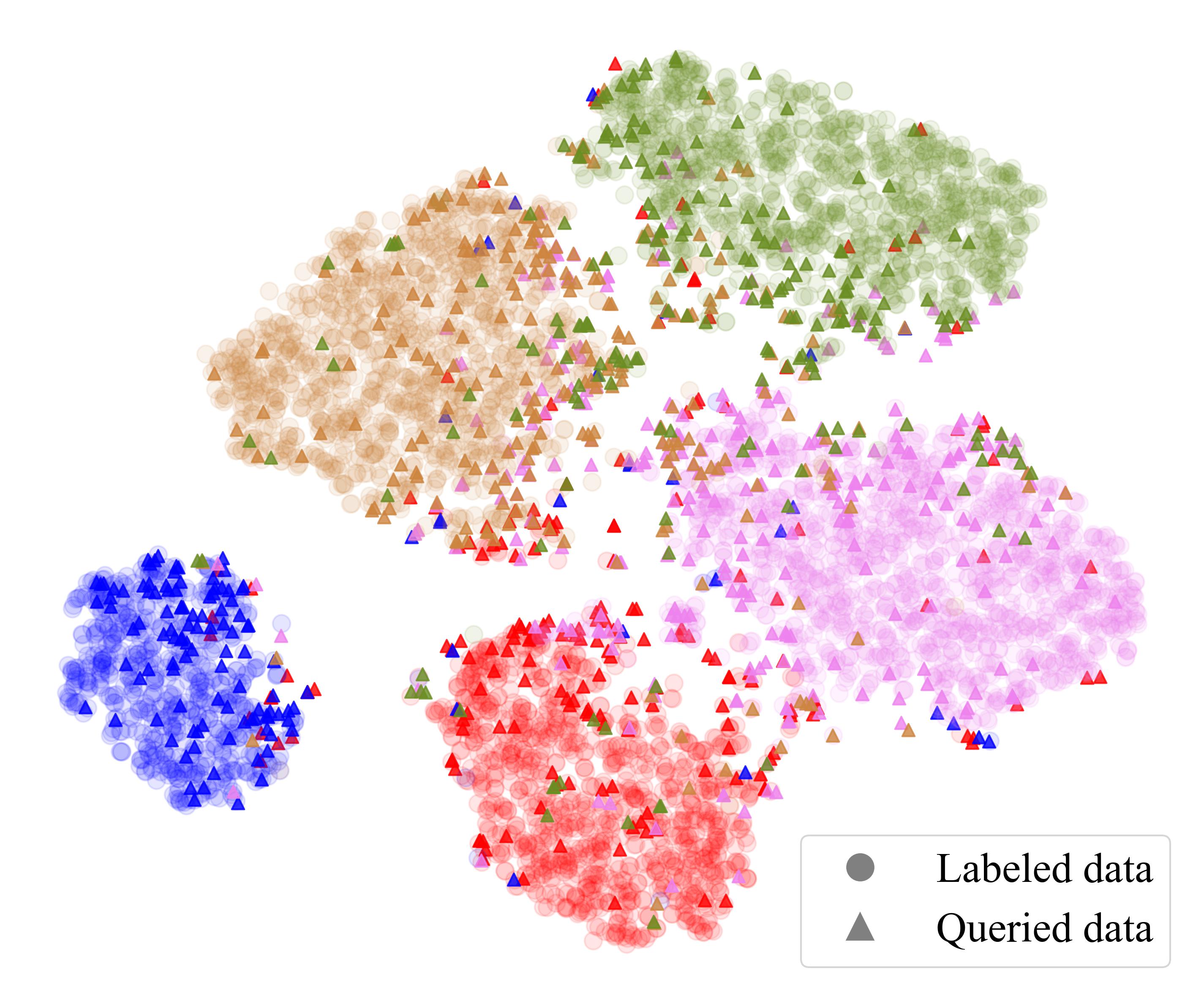}
		\caption{BUAL (ours)}
		\label{fig.tsne_b}
	\end{subfigure}
	\caption{The t-SNE feature visualization of data from one query and labeled pool on CIFAR-10 with an openness ratio of 0.5.}
	\label{fig.tsne}
\end{figure}

Figure \ref{fig.querys} shows the average recognition rate of known class examples across queries and openness ratios. We can observe that traditional AL methods all perform poorly due to their inability to recognize known class examples from those with high uncertainty and/or representative. As an OSR method, DIAS shows only marginal improvement compared to the traditional AL methods, unable to fully exploit its exceptional performance with limited training data. The recognition ability of CCAL decreases slightly with the increase of dataset categories compared to our method. It is noted that LfOSA achieved the highest query precision on all datasets. However, combining the results in Table \ref{table:acc}, we can find that the performance of models trained by LfOSA is not satisfactory compared to other methods. To explain why, we visualize the queried examples' feature representation and labeled examples' feature representation for our method and LfOSA in Figure \ref{fig.tsne}. We can observe that the features of examples queried by LfOSA are highly overlapping with the labeled ones, which are often examples that the model has already mastered and cannot provide an effective bonus for model training. On the contrary, the sample features queried by our method have little overlap and are more distributed in low-density regions, consistent with the goal of AL. These results validate the effectiveness of our approach in querying more informative examples of known classes and maintaining a high recognition rate.

\begin{table}[tb]
\centering
\caption{Final accuracy of each component in Equation \ref{eq11} on CIFAR-10, CIFAR-100, and Tiny-Imagenet with an openness ratio of 0.6.}
\label{ab}
\addtolength{\tabcolsep}{5pt}
\small
     \centering
     {\begin{tabular}{c|c|c|c|c|c} 
\toprule \toprule
\multicolumn{1}{l|}{\diagbox{Dataset}{Method}} & $\boldsymbol{unc_p}$ & $\boldsymbol{unc_n}$ &  $\boldsymbol{w/o}$ $\boldsymbol{w}$ & $\boldsymbol{w/o}$ $\boldsymbol{f_{aux}}$     & B-LC                       \\ 
\midrule
CIFAR-10  & 87.5  & 89.4  & 90.8 & 91.3 & \textbf{92.5} \\
\midrule
CIFAR-100  & 54.0  & 63.5  & 62.4 & 65.0  & \textbf{67.5} \\
\midrule
Tiny-Imagenet  & 48.4  & 52.3  & 52.0 & 53.0  & \textbf{54.7} \\
\bottomrule \bottomrule
\end{tabular}}
\end{table}

\subsection{Ablation Study}

The ablation study is conducted on CIFAR-10 with an openness ratio of 0.6 to validate the effectiveness of each component in our proposed query strategy (Equation \ref{eq11}). The final round accuracy is shown in Table \ref{ab}. Here, $\boldsymbol{unc\_n}$ and $\boldsymbol{unc\_p}$ indicate that only $unc_n$ and $unc_p$ is adopted for active sampling, respectively. $\boldsymbol{w/o}$ $\boldsymbol{w}$ denotes the removal of all balancing factors, \ie, both $f_{aux}$ and $r$. $\boldsymbol{w/o}$ $\boldsymbol{f_{aux}}$ means only removing the local balancing factor $f_{aux}$.

We can observe that removing any of the components leads to performance degradation. Although $\boldsymbol{unc_n}$ and $\boldsymbol{unc_p}$ are significantly less effective than B-LC, $\boldsymbol{unc_n}$ has a substantial improvement compared to $\boldsymbol{unc_p}$ due to its ability to distinguish informative examples of known classes from examples of unknown classes. The direct combination of $unc_n$ and $unc_p$, \ie, $\boldsymbol{w/o}$ $\boldsymbol{w}$, works better than $\boldsymbol{unc\_n}$, as models trained by RLNL may produce oscillate output and thus the uncertainty obtained in $\boldsymbol{unc\_n}$ is not necessarily accurate. By adding the dynamic global balancing factor $r$, $\boldsymbol{w/o}$ $\boldsymbol{f_{aux}}$ achieves a better performance. However, it still falls short in comparison to B-LC, which validates the effectiveness of local balancing factor $f_{aux}$. These results further corroborate the soundness of our strategy design.

\section{Conclusion}

In this paper, we successfully expand the existing uncertainty-based active learning methods to complex and ever-changing open-set scenarios by proposing a Bidirectional uncertainty-based Active Learning (BUAL) framework. On one hand, to achieve the goal of distinguishing known and unknown class examples with high uncertainty, we propose a simple but effective Random Label Negative Learning (RLNL) method for pushing unknown and known class examples toward the high and low confidence regions respectively. On the other hand, to better measure sample uncertainty, we propose a Bidirectional Uncertainty (BU) sampling strategy by dynamically fusing the sample uncertainty obtained from positive learning and negative learning. The dynamic balancing factors in it can ensure that the strategy is effective under various openness ratios. Extensive experimental results show that the model trained with BUAL can achieve state-of-the-art performance under various open-set scenarios.

\section*{Acknowledgements}

This work was supported by the Natural Science Foundation of Jiangsu Province of China (BK20222012,
BK20211517), the National Key R\&D Program of China (2020AAA0107000), and NSFC (62222605).

%
%
\bibliographystyle{splncs04}
\bibliography{main}
\end{document}